\documentclass[letterpaper, 10 pt, conference]{ieeeconf}  

\IEEEoverridecommandlockouts                              

\usepackage{graphicx}
\usepackage{amsmath}
\usepackage[backend=bibtex,style=ieee,natbib=true]{biblatex}

\usepackage[colorlinks=true,bookmarks=true]{hyperref}

\usepackage{xcolor}

\makeatletter
\newcommand{\figcaption}[1]{\def\@captype{figure}\caption{#1}}
\newcommand{\tblcaption}[1]{\def\@captype{table}\caption{#1}}
\makeatother

\usepackage[hang,small,bf]{caption}
\usepackage[subrefformat=parens]{subcaption}
\usepackage{array}
\usepackage{booktabs}
\usepackage{multirow}
\usepackage{caption}
\usepackage[flushleft]{threeparttable}
\usepackage{float}
\usepackage{soul}
\sethlcolor{red}

\usepackage{svg}

\usepackage{amssymb}

\title{\LARGE \bf
Collective Intelligence for 2D Push Manipulations with Mobile Robots
}

\author{So Kuroki$^{1}$, Tatsuya Matsushima$^{1}$, Jumpei Arima$^{2}$, Hiroki Furuta$^{1}$, \\
Yutaka Matsuo$^{1}$, Shixiang Shane Gu$^{3}$, and Yujin Tang$^{3}$
\thanks{$^{1}$The University of Tokyo, Japan}%
\thanks{$^{2}$Matsuo Institute, Japan}%
\thanks{$^{3}$Google Research, Brain Team}%
\thanks{Email: so.kuroki1931@gmail.com}%
\thanks{The source code:  https://github.com/Kuroki1931/CIOM}%
}

\begin{document}

\maketitle
\thispagestyle{empty}
\pagestyle{empty}

\begin{abstract}

While natural systems often present collective intelligence that allows them to self-organize and adapt to changes, the equivalent is missing in most artificial systems. We explore the possibility of such a system in the context of cooperative 2D push manipulations using mobile robots. Although conventional works demonstrate potential solutions for the problem in restricted settings, they have computational and learning difficulties. More importantly, these systems do not possess the ability to adapt when facing environmental changes. In this work, we show that by distilling a planner derived from a differentiable soft-body physics simulator into an attention-based neural network, our multi-robot push manipulation system achieves better performance than baselines. In addition, our system also generalizes to configurations not seen during training and is able to adapt toward task completions when external turbulence and environmental changes are applied. Supplementary videos can be found on our project website:
 \url{https://sites.google.com/view/ciom/home}.

\end{abstract}

\section{Introduction}

Collective intelligence plays an important role in natural systems, where various life forms often demonstrate self-organized and adaptive behaviors that are indispensable for them to survive the changing environment and achieve what is beyond the capabilities of any individual among the same species. For example, army ants are able to use their bodies to form a bridge that adapts to the width of a gap for the others to pass by~\cite{janel2011}. And a school of fish can change its formation during motion quickly to get rid of a predator or to hunt so that their survival rate is maximized~\cite{herbert2011inferring}. If these collective behaviors have allowed the species to survive the cruelty of evolution, can we build artificial systems that are based on the concept of collective intelligence and benefit from it too? In this work, we explore the possibility in object push manipulations by a group of cooperative mobile robots. 

As more mobile robots become readily available and move out from factories to diverse real-world settings, we see potential extensions of their applications in daily scenarios such as room cleaning, where object manipulation through pushing is the common fundamental building block of the upper level skills. Prior research mainly works on push manipulation for rigid objects~\cite{mason1982manipulator, mason1986scope, stuber2018feature, wang2002object}, when facing deformable objects, the drastic change of the underlying dynamics can make these methods struggle. A robot's capability to handle a variety of objects, especially in an indoor environment with both rigid (e.g., boxes, books) and deformable objects (e.g., cables and cloths), is indispensable in the real-world applications that involve object transport and efficient path planning~\cite{frank2014learning}. In addition, recent works have been focusing on achieving object manipulation systems with multiple mobile robots. Compared with a single robot setting, such systems are more valuable for several reasons~\cite{cao1997cooperative}. For instance, while pushing a single rigid object at a time involves straight-forward kinematics, pushing a deformable object presents unique challenges as dynamics becomes more complex and dexterity demands multi-robot coordination. Moreover, building several simple robots can be more cost effective and fault tolerant than having a single powerful robot.

\begin{figure}[!t]
    \centering
    \includegraphics[width=0.4\textwidth]{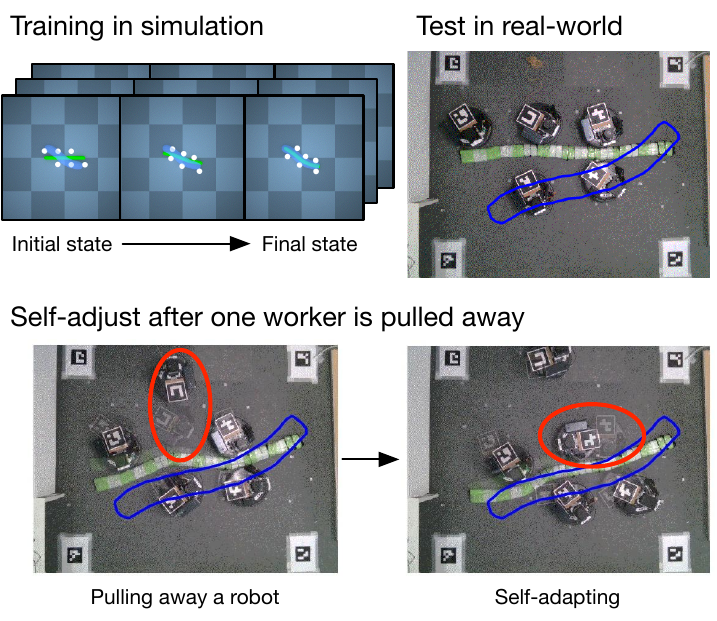}
    \caption{Our robots trained in simulation (top left) generalize to the real world tests (top right). They even learned to adapt to the change when one of the robots was pulled away during task execution (bottom left). In this case, a neighboring robot quickly compensated for the missing robot and worked toward the completion of the task (bottom right). In all figures, the blue lines describe the goal poses for the rope. In the bottom row, we highlight the robots pertaining to the turbulence and the adaptation, and superpose the state from the previous time step translucently to illustrate the changes.}
    \label{fig:cover}
    \vskip -0.3in
\end{figure}

Crucially, a multi-robot push manipulation system allows dexterity. By commanding the robots to the appropriate contact points, and adjusting their relative distances between each other based on the target object's state, we have more flexible wrench spaces. Here, achieving system level coordination has been the main goal for many research works, and object manipulation using cooperative mobile robots can be viewed as a specific case of multi-robot coordination~\cite{feng2020overview}.
Early methods often assign leaders and followers roles to the robots so that the complex task is factored and heuristic solutions can be easily incorporated~\cite{brown1995pusher,wang2007decentralized}.
In this scheme, although the leaders usually bear more computational burdens, the problem is approached in a structured way and the overall complexity is reduced.
\cite{montemayor2005decentralized,farivarnejad2016decentralized} explored decentralized multi-robot motion control in a simplified environment where robots are assumed to be rigidly attached to the load in a plane. Alternatively, by treating the cooperative load transport problem as a formation control problem, \cite{bai2010cooperative} allows all the agents to achieve a desired constant velocity during motions in cooperative transportation tasks.
With the development of physics simulators~\cite{hu2019difftaichi,heiden2021disect,freeman2021brax,makoviychuk2021isaac} and faster hardware accelerators, we are witnessing better performance brought by trajectory optimization/generation based methods and learning based methods these days. For example, \cite{burget2016bi} obtained paths for mobile manipulators with methods based on Rapidly-exploring Random Tree (RRT)~\cite{lavalle1998rapidly}, and \cite{honerkamp2021learning} uses reinforcement learning to generate the trajectories for mobile manipulation actions.
These works demonstrate the potential of trajectory optimization and learning based methods with the help of modern physics engines. On the other hand, they may require a prohibitive amount of computing at each decision step and/or have difficulties in learning the task due to the long task horizons. Furthermore, these systems do not present the collective intelligence we mentioned earlier in the natural system that would allow them to adapt when facing environmental changes and system failures.

In this paper, we show that by distilling a planner derived from a differentiable soft-body physics simulator into an attention~\cite{waswani2017attention} based neural network, our multi-robot manipulation system can achieve better performance than both trajectory optimization and reinforcement learning methods. We train our multi-robot system in a simulator, and demonstrate that it generalizes well to scenarios with physics configurations and numbers of robots that are different from what were given during the training. After training, we zero-shot transfer the learned policy to the real world and the system is able to keep most of its performance. Finally, our robots are able to self-adjust in the face of environmental changes. For example, we pulled away one of the robots during task execution, and observed a neighboring robot proactively adjusted its position to compensate for the loss of the co-worker (see Figure~\ref{fig:cover}). This cooperative and self-adaptive behavior demonstrates the emergent collective intelligence unseen in conventional systems. In terms of real-world applications, we envision a multi-robot indoor cleaning system (e.g., inside a shopping mall), where a subset of the robots move the objects to the pre-defined positions/poses so that the paths are cleared for the others.

\section{Related Work}

\subsection{Multi-Robot Push Manipulations}

In push manipulation, it is important to model the object dynamics with collisions and friction with respect to the manipulator's action~\cite{stuber2020let}.
For dynamics modeling, early works studied under the quasi-static assumption~\cite{mason1982manipulator, mason1986scope} that approximates the equations of motion in the horizontal direction, assuming that the vertical scale of the associated object is sufficiently small relative to the horizontal. After that, we see a surge of data-driven methods applied for modeling dynamics. To avoid explicit estimations of physics parameters, data-efficient forward model construction methods have been proposed to learn the dynamics directly~\cite{stuber2018feature,byravan2017se3}.
Recently, the drastic improvement of modern physics engines in terms of both precision and speed has allowed researchers to develop planners more efficiently~\cite{zito2012two, king2016robust}. In this work, we also rely on a modern differentiable physics engine to provide us the related dynamic models.

The latest works have also put dexterity at its core~\cite{dafle2014extrinsic,zhou2022learning}. In the context of push manipulation, we argue that a multi-robot system is more suitable in many scenarios because it allows much more flexibility in wrench spaces. Pioneering works have demonstrated the potential of multi-robot manipulation systems in various, albeit restricted, settings. For example, relying on predefined motion primitives, \cite{nilles2021information} designed robust motion strategies for micro-manipulations through boundary interactions.
In~\cite{kim2015new}, the authors studied the dynamics of a multi-robot object clustering system that focus on boundaries between objects and the scaled compactness. \cite{amato2016policy} introduced a planning algorithm that searches over policies represented as finite-state controllers for a cooperative multi-robot system under uncertainties. The algorithm generates controllers for a heterogeneous robot team that collectively maximizes the team utility in a bar-tendering task. We aim to develop a multi-robot push manipulation system of homogeneous robots with the equivalent flexibility but fewer restrictions.

\subsection{Collective Intelligence in Machine Learning and Robotics}
Various works incorporated collective intelligence concepts into artificial systems and have proved the merits in doing so~\cite{ha2022collective,gauci2018programmable,slavkov2018morphogenesis}. For example, it has been shown that systems with neural cellular automata implementations can not only model complex 2D/3D patterns but also demonstrate strong robustness to external turbulences through regeneration~\cite{mordvintsev2020growing,sudhakaran2021growing}. In multi-agent setups, complex collective behaviors emerge automatically to allow efficient cooperative/competitive strategies~\cite{zheng2018magent}, coordinated modular locomotion controllers~\cite{huang2020one} and flexible structure designs~\cite{pathak2019learning,gonzalez2022influencing}.
In addition to these simulated environments, roboticists have also reproduced well known collective behaviors found in natural systems in robotic systems. For instance, \cite{mccreery2022hysteresis} studies the principle behind the army ants' bridge construction behavior and identifies the control mechanism emerged from each individual’s decisions. Based on this, \cite{malley2020eciton} then designed self-assemble robots that climb over each other to form amorphous structures. Similarly, fish schooling behaviors are also reproduced in underwater robots that present natural and self-organized flocking states~\cite{gauci2014self,berlinger2021implicit}. In the space of cooperative manipulation, multiple works have learned to control swarm robotics. In the task of pushing a horizontal box, pushing with multiple robots completed the task more efficiently than with a single robot~\cite{shen2022deep,mataric1995cooperative}. Following these works, we wish to encapsulate the self-organizing capability in our system so that it is robust and more suitable for real-world challenges. 


\subsection{Differentiable Physics Engines}
Differentiable physics engines leverage differentiable programming for physics simulations.
Relying on automatic differentiation, differentiable simulators propagate gradients through dynamics and can be utilized for parameter estimation~\citep{ma2022risp}, gradient-based trajectory optimization~\citep{huang2021plasticinelab,hu2019difftaichi}, and policy learning~\citep{freeman2021brax,mora2021pods}.
While most differentiable simulators are tailored for rigid-body physics~\cite{freeman2021brax}, we see recent simulators implementing particle dynamics~\citep{hu2019difftaichi,warp2022} which, in the context of object manipulation, unlocks the learning of non-rigid object manipulations. For instance, \cite{qi2022learning} conducted sim2real transfer of dough manipulation with imitation learning with the help of such simulators.
Our method is based on PlasticineLab (PL)~\citep{huang2021plasticinelab}, a differentiable simulator benchmark using DiffTaichi~\citep{hu2019difftaichi} that enables gradient-based trajectory optimization through particle dynamics. As we will show later, after our proper data interface design, a gradient-based trajectory planner achieves superior performance compared with baselines.

\subsection{Multi-Head Self-Attention}
We encapsulate ``self-organizing'' capabilities in our control policy via a multi-head self-attention mechanism.
A (self-)attention module enables a network to develop fast weights that adapt quickly based on the input data. Since its release, we have witnessed successful applications of attention in many areas~\cite{waswani2017attention,lee2019set,liu2021swin,chen2021decision,tang2021sensory}. In general, a self-attention module can be mathematically described as:
\begin{equation}
Y=\mathrm{softmax}\big{(}\frac{1}{\sqrt{d_{in}}}(XW_q)(XW_k)^\intercal\big{)}(XW_v)
\end{equation}
where $X \in \mathcal{R}^{N \times d_{\mathrm{in}}}$ is the input data, $Y \in \mathcal{R}^{N \times d_{v}}$ is the output, $W_q, W_k \in \mathcal{R}^{d_{\mathrm{in}} \times d_{k}}$ and $W_v \in \mathcal{R}^{d_{\mathrm{in}} \times d_{v}}$ are learnable projection matrices. If there are multiple sets of projection matrices, we can concatenate the outputs and have $Y=[Y_1, \cdots, Y_m]$, we call such a setting multi-head self-attention where $m$ is the number of heads.

\begin{figure*}[t]
    \begin{center}
    \includegraphics[width=0.75\textwidth]{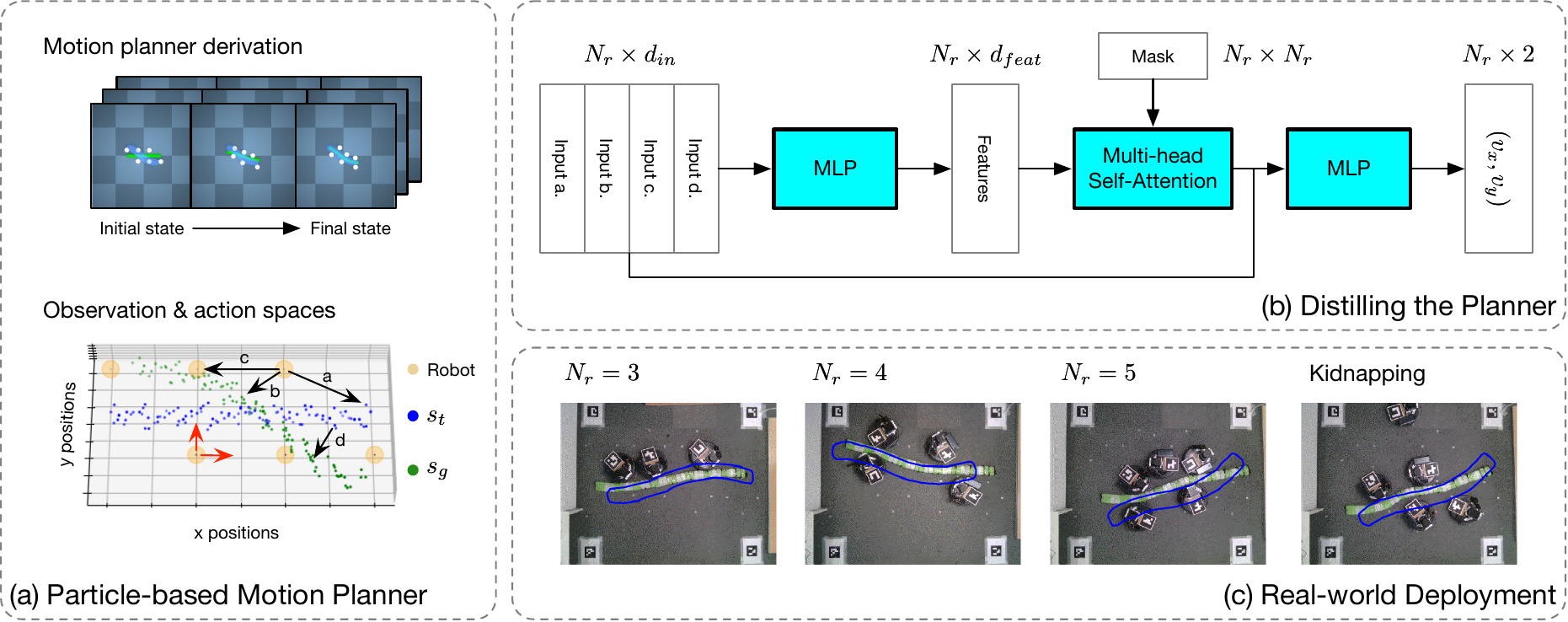}
    \caption{An overview of our method. (a) In this phase, we focus on designing the data interfaces and evaluating a derived gradient-based motion planner. $s_t$ and $s_g$ represent the current and the goal states of the target object, a-d are four input components considered by each robot (see Section~\ref{sec:distillation} for details). (b) We collect data from the planner and distill the behavior into an Attention-based neural network policy. The data dimension $d_{in}=d_a+d_b+d_c+d_d$ is an aggregation of the four input parts described in Section~\ref{sec:distillation}. (c) We zero-shot transfer the distilled policy onto real robots and conduct tests, including settings that were not seen during training.}
    \label{fig:overview}
    \end{center}
    \vskip -0.25in
\end{figure*}

\section{Proposed Method}



\subsection{Problem Statement}
We design and study our system in the settings where multiple mobile robots (e.g., Roomba) are required to push and manipulate a target body object (e.g., a rope) to a user defined goal pose. Our key assumption is the availability of full top-down views of the workspace, which allows us to extract the pose information of both the robots and the target object in centralized data collation. In addition, we also assume the interaction between a robot and the target object to be one-way force coupling. We define $r(s_t)$ as a measurement of the system's performance at time step $t$:
\begin{equation}
r(s_t) = \mathrm{max}\left(\dfrac{{f({s}_{t}, {s}_{g})} - {f({s}_{0}, {s}_{g})}}{\mathrm{max}(1 - {f({s}_{0}, {s}_{g}), \epsilon)}}, 0\right)
\end{equation}
where $s_t$ and $s_g$ represent the current and the user defined goal state of the target object respectively, $f(s_i, s_j) \in [0, 1]$ measures the similarity between $s_i$ and $s_j$ ($f(s_i, s_j)=1, \forall i=j$), and $\epsilon$ is a small positive number to prevent the zero division error. Following~\cite{huang2021plasticinelab}, $s_t$ and $s_g$ are collections of particles on the object, we map these particles into a common 3D grid space covering the workspace, and use intersection over union (IoU) as the similarity function $f$.

\subsection{Gradient-based Motion Planning}

Our method is based on the particle system dynamics provided in PL.
In the first step of our workflow, we define the data interfaces and evaluate a gradient-based motion planner (see Figure~\ref{fig:overview} (a) top). Specifically, we model the target object as a particle system and extract $N_p$ particles to form data of shape $N_p \times d_p$ that represents the target object's current and goal states $s_t$ and $s_g$. Concretely, we first define a cubic function to specify the goal pose. We then extract equally spaced points along the function's curve to form a base set of particles. These particles are in the center of the cross sections along the rope. By adding equally spaced particles on those cross sections, we are able to get a ``wire frame'' of the object. We further sample uniformly inside the wire frame to construct $s_g$, once these particles are determined, they are fixed throughout the task. Particles in $s_t$ are identified in a similar fashion: we use color matching to detect the target object from the global camera, we then use regression to find a function that describes the current pose, after that we apply the same steps as we did in forming $s_g$. $N_p=2328$ is the number of particles and $d_p=6$ (the positions and linear velocities in the 3D space) is the information dimension of each particle. In 2D push manipulations $d_p=4$ suffices, but we keep a uniform interface for more complex tasks in the future. PL treats the mobile robots as rigid body manipulators, and we therefore encapsulate their states in a $N_r \times d_r$ matrix, where $N_r$ is the number of robots and $d_r=5$ contains each robot's positions and linear velocities (we ignore $v_z$ because we assume the linear velocities in the XY plane alone is sufficient to solve the problem, see Figure~\ref{fig:overview} (a) bottom).

We define Equation~\ref{eq:train_loss} as our loss function at each step:
\begin{equation}\label{eq:train_loss}
L_t=c_1 L_{\mathrm{mass}} + c_2 L_{\mathrm{dist}} + c_3 L_{\mathrm{grasp}}
\end{equation}
where $L_{\mathrm{mass}}$ measures the L1 distance between $s_t$ and $s_g$, $L_{\mathrm{dist}}$ is the dot product of the signed distance field of $s_t$ and $s_g$, and $L_{\mathrm{grasp}}$ is the sum of distances between the robots and the object.
Our method employs a differentiable physics engine, and because Equation~\ref{eq:train_loss} is also differentiable, it is straightforward to derive a gradient-based motion planner (GMP) for the robots. In our experiments, we used Adam~\cite{kingma2014adam} to optimize randomly initialized velocity commands for each robot, and we have summarized the parameters in Table~\ref{tab:hps}. As we will show shortly, although GMP's performance is better than the baselines, querying it at every time step is less efficient. Moreover, it is not straightforward for such a planner to adapt to environmental changes (e.g., changes of physics configurations or the number of robots). We therefore propose to distill this GMP into an attention-based neural network policy that overcomes these shortcomings while maintaining the performance.


\subsection{Distilling the Planner}
\label{sec:distillation}

Our self-attention based policy is depicted in Figure~\ref{fig:overview} (b), the two Multi-layer Perceptrons (MLPs) are single-layered networks (i.e., no hidden layers). To contain the policy size, we uniformly down-sample from the $N_p=2328$ particles to $N_{p'}=102$ and fix these particles throughout the episode. The input to our policy is a matrix of shape $N_r \times d_{in}$, where the rows describe the sensory information collected by each of the $N_r$ robots. In our design, each robot's sensory information is a concatenation of 4 parts (i.e., $d_{in}=d_{a} + d_{b} + d_{c} + d_{d}$): (a) the positions of the particles of the target object in $s_t$ in the robot's frame ($d_{a}=3N_{p'}$); (b) the positions of the particles of the target object in $s_g$ in the robot's frame ($d_{b}=3N_{p'}$); (c) the relative position of its nearest neighbor robot ($d_{c}=3$); and (d) the vectors from the object particles in $s_t$ to those in $s_{g}$ ($d_{d}=3N_{p'}$). Although this last component is common to all robots and can be input into the policy at a later stage, we choose to combine with each robot's sensory data for convenience. Receiving the input, our policy transforms the data into features of shape $N_r \times d_{feat}$ via an MLP, and then several layers of multi-head self-attentions are applied. A visibility mask is used here to make sure each robot has access to only the data from itself and its nearest neighboring robot. Intuitively, this attention module allows each robot to aggregate its own knowledge about the world with those from its nearest neighbor. Finally, we concatenate the raw sensory data with the attention layers' outputs and use another MLP to output the velocity commands $(v_x^{(i)}, v_y^{(i)}), i=1 \cdots N_r$.

Notice that the number of robots $N_r$ is treated as the batch dimension commonly seen in deep learning, and the change of which (in terms of both order and size) would not interfere with our policy's data processing procedure. This characteristic allows more flexible deployment of the policy when facing different numbers of robots that are seen during training. Moreover, the visibility mask forces the policy to get accustomed to real-world settings where robots can only see a subset of their fellow workers due to communication restrictions. To train the policy network, we collect a dataset $D$ of size $2125$ from the GMP with $N_r=6$ robots trained on 949 goal state configurations and three task horizons $T \in \{40, 70, 100\}$. In policy distillation, we predict the GMP's actions $a_t=(v_x, v_y)_t,t=0, \cdots, T-1$ and use mean squared error as the loss.

\subsection{Real-world Policy Deployment}
\begin{figure}[t]
    \begin{center}
    \includegraphics[width=0.55\linewidth]{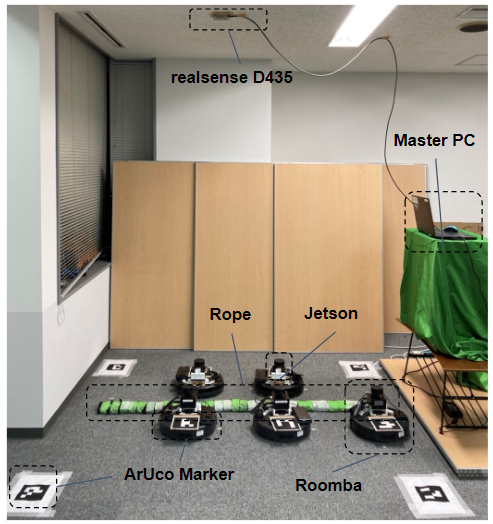}
    \caption{Illustration of our real-world experimental setup. We define the arena area and acquire the positions of the robots via the ArUco markers on the floor and on the Roombas.}
    \label{fig:setup}
    \end{center}
    \vskip -0.23in
\end{figure}

Although our policy is only trained with data collected from simulations, we deploy and test the policy in the real world with various configurations (see Figure~\ref{fig:overview} (c) for sample configurations) without any fine-tuning. Figure~\ref{fig:setup} gives our hardware setups. We mount a camera on the ceiling to capture global views, and then analyze the point cloud from the camera to extract the parts corresponding to the target object by matching the colors. Each robot bears a marker, from which its positions in the environment are detected. A central server is responsible for collecting sensory data, executing the policy and distributing the velocity commands to all robots. To avoid jerky motions, we smooth the actions by averaging the latest $H$ policy outputs.

\section{EXPERIMENTS}
We answer the following questions via experiments.
\begin{itemize}
    \item How does the GMP compare with other methods?
    \item Can we distill the GMP into a neural network policy and keep the performance?
    \item And if so, will it adapt to environmental changes and present collectively intelligent behaviors?
\end{itemize}

\subsection{Experimental Setup}
\label{sec:ex_setup}

We use a rope as our target object and command a group of Roombas to push manipulate the rope into user defined poses. 
In policy distillation, we set $d_{feat}=128$ and train on the collected dataset until the error converges. In tests, we use $N_r \in \{3, 4, 5\}$ robots. For each test, we fix the initial state of the robots and the rope but, depending on $N_r$, sample the goal state from 50 (when test in sim) and 4 (when test in real) predefined configurations. In generalization tests, while the physics configurations during training are fixed, they are randomly sampled from predefined ranges in tests (see Table~\ref{tab:hps} Physics Configurations). In the real-world deployment, each Roomba has an ArUco marker so that the global camera can detect their positions. All data communication and command distributions are conducted via ROS. The robots' control frequency is set to 10Hz and the action smoothing window size is $H=5$.
Although the sensory data are inevitably noisy in the real-world compared with those in the simulator, our policy is still able to accomplish the tasks, demonstrating the robustness of our method.


We introduce several baseline methods to compare the performances. Concretely, we trained a Proximal Policy Optimization (PPO)~\cite{schulman2017proximal} agent with an MLP policy (two hidden layers with 64 hidden units each), and a Model Predictive Path Integral (MPPI) planner on the same task (see Table~\ref{tab:hps} for detailed parameters). The MPPI baseline also relies on the dynamics provided by PL, where the particle dynamics follows material point method and the soft body collision with rigid bodies utilizes grid-based contact treatment with Coulomb frictions. Both MPPI and PPO optimize the same loss function as GMP does. We also compare the performance of the distilled policy with the teacher policy (i.e., the GMP) to show that our method is able to maintain the performance. Finally, to demonstrate the importance of the self-attention module for generalization, we distilled the planner into an MLP policy for comparison.

\begin{table}[!t]
\centering
\setlength{\tabcolsep}{4pt}
\caption{Summary of hyper-parameters.}
\begin{tabular}{lllll}

\multicolumn{2}{c}{GMP}                  &  & \multicolumn{2}{c}{Physics Configurations}         \\ \cline{1-2} \cline{4-5}
\textbf{Parameter}    & \textbf{Value}   &  & \textbf{Parameter} & \textbf{Train (Test Range)}   \\ \cline{1-2} \cline{4-5}
density loss ($c_1$)  & $500$            &  & Friction           & $1.5$ ($[1.0, 2.5]$)          \\
sdf loss ($c_2$)      & $500$            &  & Rope yield stress  & $30$ ($[15, 45]$)             \\
contact loss ($c_3$)  & $1$              &  & Velocity limit     & $0.015$ ($[0.005, 0.02]$)     \\
learning rate         & $0.1$            &  &                    & (for two robots)              \\
optimizer             & Adam             &  & Robot radius       & $0.02$ ($[0.02, 0.035]$)      \\ \cline{1-2} \cline{4-5} 

\\

\multicolumn{2}{c}{PPO}             &  & \multicolumn{2}{c}{MPPI}            \\ \cline{1-2} \cline{4-5} 
\textbf{Parameter} & \textbf{Value} &  & \textbf{Parameter} & \textbf{Value} \\ \cline{1-2} \cline{4-5}
learning rate      & $3e-4$         &  & \#samples          & 1200           \\
entropy coef       & $0.01$         &  & horizon            & 100            \\
value loss coef    & $0.5$          &  & sampling stage     & 30             \\
batch size         & $32$           &  & noise mean         & 0              \\
epochs             & $10$           &  & noise std          & 1              \\ \cline{1-2} \cline{4-5} 
\end{tabular}
\label{tab:hps}
\vskip -0.1in
\end{table}

\subsection{Evaluating the Gradient-based Motion Planner}

\begin{figure}[t]
    \begin{center}
    \includegraphics[width=0.4\textwidth]{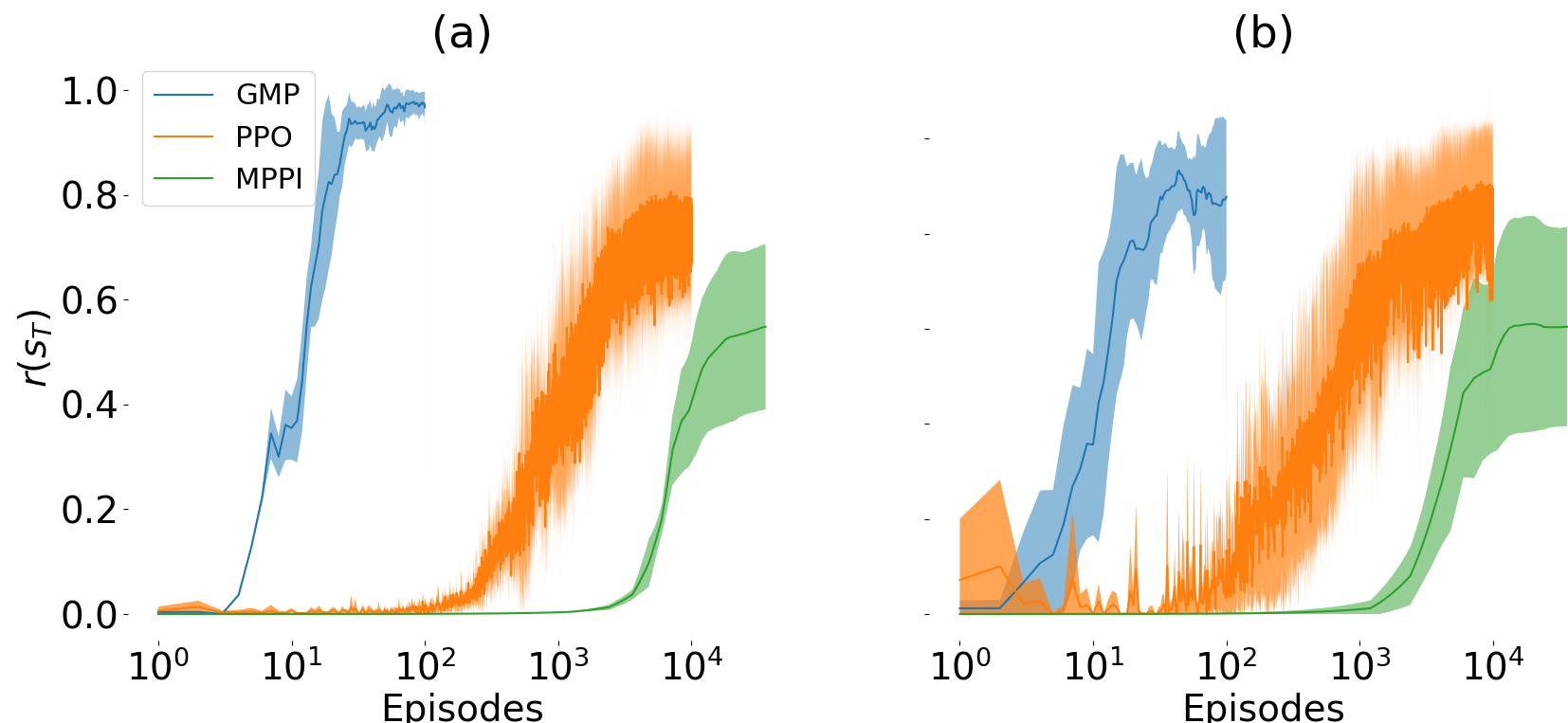}
    \caption{Comparison of $r(s_T)$ in rope manipulation. For comparison, we fixed the number of the robot at $N_r=6$. (a) The mean reward and standard deviation for each episode of 5 seeds with the same task. (b) The mean reward and standard deviation for each episode of 5 tasks with the same seed.}
    \label{fig:compare_reward}
    \end{center}
    \vskip -0.2in
\end{figure}
\begin{figure}[t]
    \begin{center}
    \includegraphics[width=0.6\linewidth]{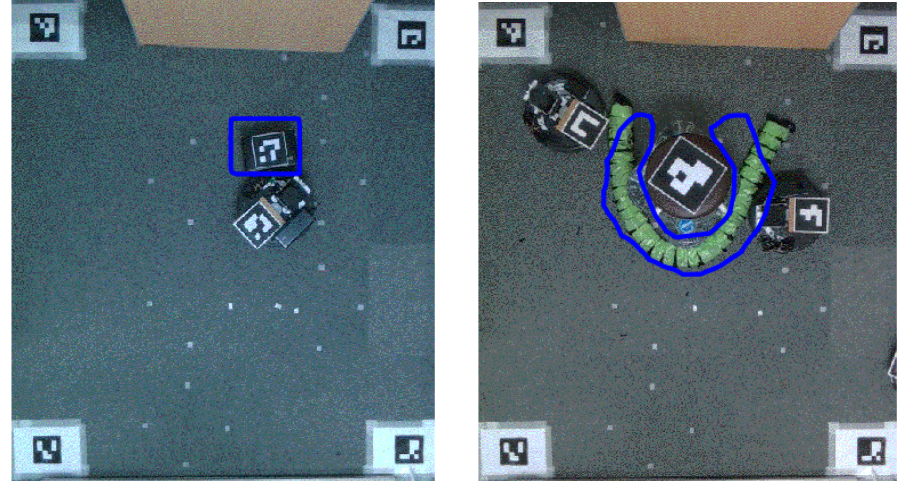}
    \caption{Applying the GMP directly in the real-world. On the left, the robot pushes a box to a specified pose configuration (the blue bounded area). On the right, two robots twist a rope into a required shape. Following\protect{\cite{huang2021plasticinelab}}, we employ a fixed environmental object as an anchor in the middle.}
    \label{fig:real_expert}
    \end{center}
    \vskip -0.2in
\end{figure}

Figure~\ref{fig:compare_reward} gives the learning curves of all methods. In these experiments, we used $N_r=6$ robots and the starting configurations of the robots and the ropes are fixed and identical for all methods. On the left, we show the results from five trials with different random seeds where a common goal configuration is sampled and fixed for all trials. On the right, we report the results from training and testing with five different goal configurations. Results show that GMP reaches a high $r(s_T)$ and a small variance within shorter episodes than the two baseline methods. MPPI shows lower performance than GMP even after sufficient sampling and aggregation iterations. This may be due to the fact that the computation time required by the sampling-based method increases with the number of robots. One possibility for improvement is to use a simplified kinematic chain to model the target object, with the trade-off of extra cost to model each object, and the accuracy may decrease for the tasks where the rope is more deflected (see Figure~\ref{fig:real_expert} right).

Confirming that GMP has better performance in simulation, we also want to investigate its sim2real gap's size so that we know what to expect when we distill the planner and evaluate our policy in the real-world. To this end, we applied the planner directly in the real-world and recorded its performances. In addition to the rope manipulation task, we also added a box pushing task (see Figure~\ref{fig:real_expert} for task snapshots). Table~\ref{tbl:real_exp} summarizes our results, where for each run, we test the planner on four different goal configurations and we report the mean and the standard deviation of $r(s_T)$. The results show a small sim2real gap and suggest the planner to be a good candidate as a teacher policy for our policy distillation. Compared with the rope pushing task, the relatively large variance in box pushing is due to the size/mass of the box and the single point of contact (there is only one robot in this task): a slightly imbalanced contact point causes the box to tilt and the score to vary.

\begin{table}[!t]
\centering
\caption{GMP sim2real gap evaluation. In each experiment, we test on four different goal states and report the mean and the standard deviation of $r(s_T)$.}
\begin{tabular}{ll}
\hline
\textbf{Task}                 & \textbf{Performance}   \\ \hline
\multirow{2}{*}{Pushing Box}  & $0.62 \pm 0.22$ (sim)  \\
                              & $0.59 \pm 0.28$ (real) \\ \hline
\multirow{2}{*}{Shaping Rope} & $0.62 \pm 0.02$ (sim)  \\
                              & $0.50 \pm 0.15$ (real) \\ \hline
\end{tabular}
\label{tbl:real_exp}
\vskip -0.1in
\end{table}

\subsection{Evaluating the Distilled Policy}

After policy distillation, we evaluated the resulting policies in the simulation of the rope manipulation task. In the first set of tests, we used $N_r=6$ robots and tested all the methods on 50 new goal states $s_g$'s. We report the mean and the standard deviation of $r(s_T)$ from each method in Table~\ref{tbl:bc_exp}, and all the distilled policies have the prefix ``BC+" for behavior cloning. Compared with the teacher policy (first row), we observe mild performance drops from the distilled policies (second and third rows). However, we expect this gap to shrink if we train with more data and/or a larger student policy.
Because an MLP policy (second row) imposes strong constraints on the input (e.g., fixed number/order of robots) that does not allow adjustment to environmental changes, we find an attention-based network policy (third row) to be more appropriate, with a trade-off in performance due to the architectural inductive bias. As we will show later, this performance trade-off unlocks much stronger generalization capabilities and is worthwhile in real-world applications. 
In the last row, we include the results from PPO that is trained from scratch. Although the method has the same MLP as our method in the second row, its performance is much lower than any of the distilled policies (statistically significant from t-tests), suggesting the crucial role played by our proposed method. In terms of time cost, our methods' average inference time are $0.015$s for the MLP policy and $0.19$s for the self-attention based policy. Compared with that of GMP's, the drastically reduced time cost demonstrates the value of our proposed method. Moreover, to show scalability to an increasing number of robots, we give the self-attention based policy dummy data for 100 robots, and the result is 0.20s. While other methods' computational time increase linearly (or worse) with of the number of robots, our method is not impacted and has the potential of real-time inference even when the number of robots increases. 

\begin{table}[!t]
\centering
\caption{Performance in the task of rope manipulation. We show the test mean and the standard deviation of $r(s_T)$ with $N_r=6$ robots on 50 new goal states. The distilled policies have ``BC +" (behavior cloning) as their prefix. The architecture of the MLP in second and fourth rows are described in Section~\ref{sec:ex_setup}.}
\begin{tabular}{lll}
\hline
\textbf{Method}            & \textbf{Performance} & \textbf{time/step(s)} \\ \hline
GMP (teacher)              & $0.74 \pm 0.28$      & $3.4   \pm 0.060 $    \\ \hline
BC + MLP                   & $0.64 \pm 0.22$      & $0.015 \pm 8.0e-4$    \\ \hline
BC + Attention (w/o delay) & $0.54 \pm 0.27$      & $0.19  \pm 0.017 $    \\ \hline
PPO + MLP                  & $0.40 \pm 0.26$      & $0.015 \pm 1.3e-4$   \\ \hline
\end{tabular}\label{tbl:bc_exp}
\vskip -0.0in
\end{table}

\begin{table}[!t]
\centering
\caption{Generalization test results. We changed several configurations to those unseen during training/data collection and report the test scores. For the experiments in the simulator, we report the mean and standard deviation of $r(s_T)$ on 50 new goal states; for the experiments in the real-world, we report the same metrics from 5 new goal states.}
\begin{tabular}{llll}
\hline
\textbf{What's changed?} & \textbf{BC+Attention} & \textbf{BC+MLP} & \textbf{PPO+MLP} \\ \hline
\multicolumn{4}{l}{(Experiments in sim)}                                              \\
Friction                 & $0.42 \pm 0.25$       & $0.50 \pm 0.27$ & $0.32 \pm 0.22$  \\
Rope yield stress        & $0.49 \pm 0.23$       & $0.58 \pm 0.25$ & $0.34 \pm 0.23$  \\
Velocity limit           & $0.50 \pm 0.25$       & $0.42 \pm 0.30$ & $0.33 \pm 0.23$  \\
Robot radius             & $0.45 \pm 0.27$       & $0.56 \pm 0.25$ & $0.34 \pm 0.24$  \\ 
\multicolumn{4}{l}{(Experiments in real)}                                             \\
$N_r=5$                  & $0.44 \pm 0.10$       & N/A             & N/A              \\
$N_r=4$                  & $0.40 \pm 0.14$       & N/A             & N/A              \\
$N_r=3$                  & $0.54 \pm 0.14$       & N/A             & N/A              \\ \hline
\end{tabular}\label{tbf:gen_exp}
\vskip -0.1in
\end{table}

\begin{figure}[!t]
    \begin{center}
    \includegraphics[width=0.4\textwidth]{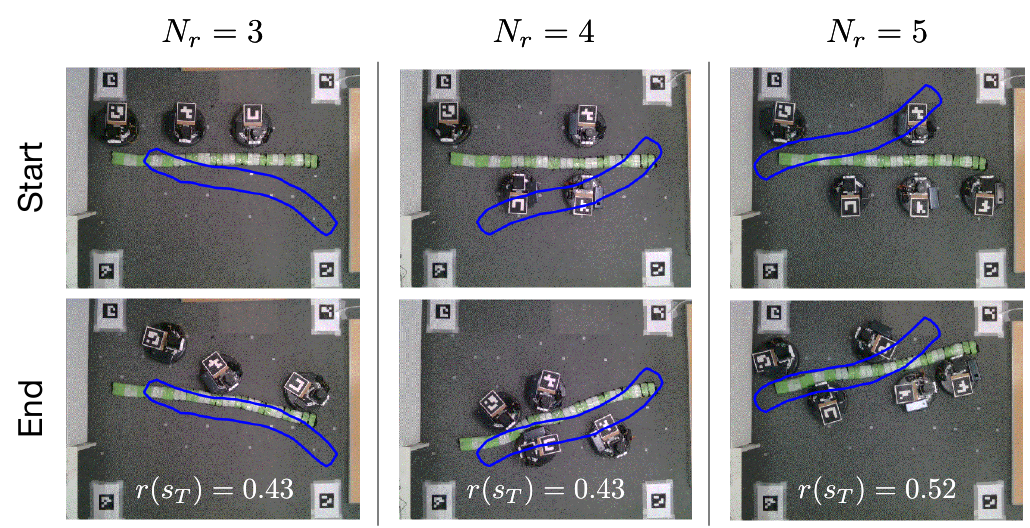}
    \caption{Test examples. We show the start and end states of the tasks in each column, and we superpose the final $r(s_T)$.}
    \label{fig:real_policy}
    \end{center}
    \vskip -0.3in
\end{figure}

\begin{figure*}[!ht]
    \begin{center}
    \includegraphics[width=0.75\textwidth]{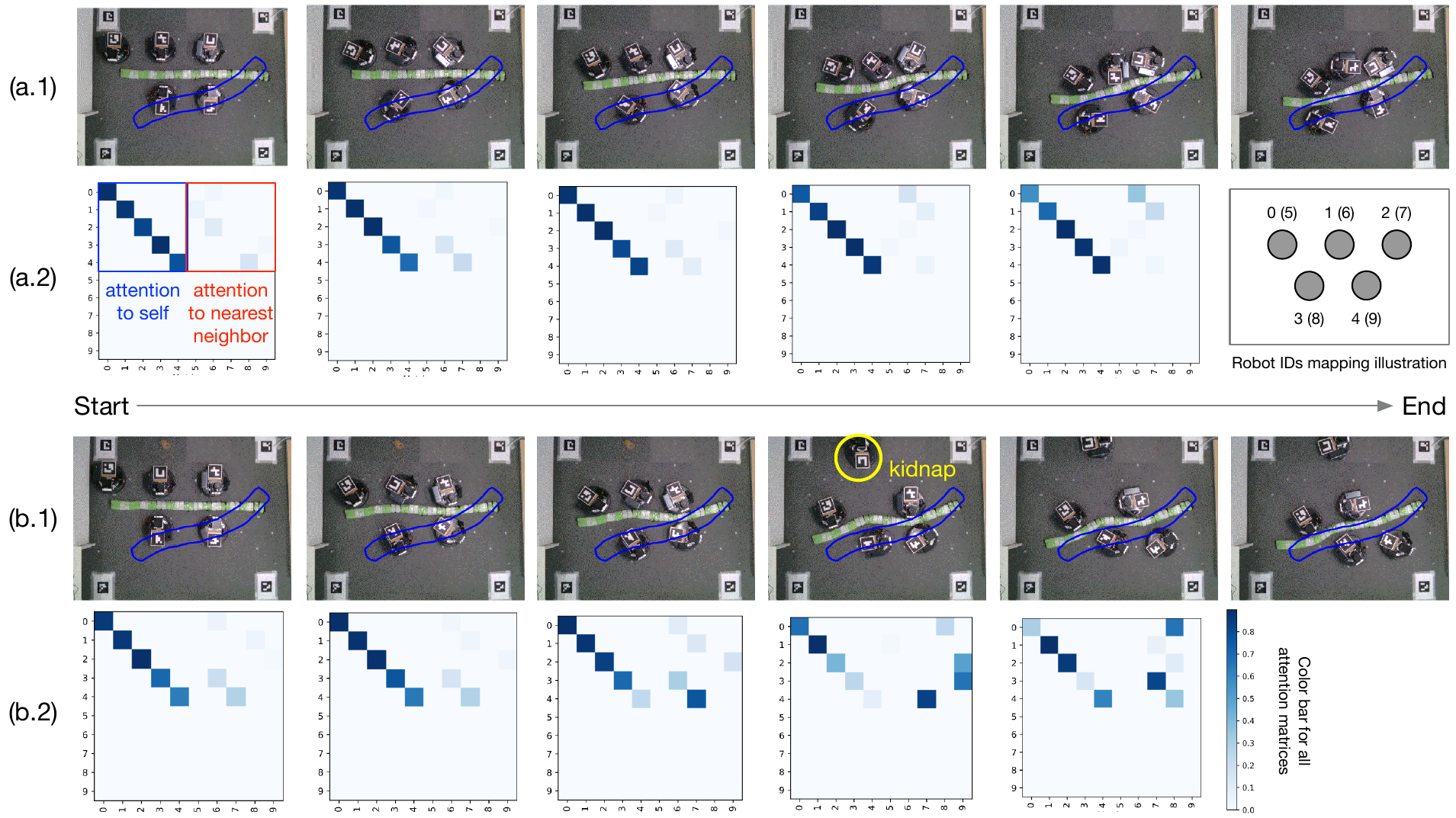}
    \caption{Behavior analysis experiments. (a.1 \& b.1) We command the robots to push manipulate the rope into a specified pose. In (b.1) we pulled away one robot in the middle of the task execution. (a.2 \& b.2) The corresponding attention matrices. In each attention matrix plot, the top left block shows how much attention each robot is paying to itself, and the upper right block shows those to the nearest neighbor. The mapping between a robot and its IDs in the attention matrices is shown in the last plot in (a.2) in the format ``ID in top left block (ID in top right block)".}
    \label{fig:self_adaptive}
    \end{center}
    \vskip -0.3in
\end{figure*}

Next, we test the distilled policy's generalization capability in both simulations and in the real world with unseen configurations. Table~\ref{tbf:gen_exp} summarizes our findings. In simulations, we vary the physics properties such as the frictions between the rope and the ground, the rope yield stress, etc (see Table~\ref{tab:hps} for all variables and values). Both BC policies are able to keep more than $80\%$ of their performances on average and are significantly better than PPO. In real-world tests, unlike an MLP policy, our attention-based network policy does not expect a fixed number/order of robots. Therefore, we are able to test with different $N_r$'s without fine-tuning or retraining (see Figure~\ref{fig:real_policy}). A major environmental change as it is, our system is able to keep $70\%$ of its performance in the worst case and is doing better than the baseline.

\subsection{Behavior Analysis}

In our last experiments, we applied even more aggressive environmental changes to the robots and analyzed the policy network's attention matrices to get insights into their behaviors. More specifically, we conducted two sets of experiments in the real world. In the first trial, we command a group of $N_r=5$ robots to push the rope into a specified pose (see Figure~\ref{fig:self_adaptive} top rows). And in the second trial, all the settings are kept identical except that we ``kidnap'' one of the robots in the middle of the task execution and let the rest of robots finish the job (see Figure~\ref{fig:self_adaptive} bottom rows). Notice that the kidnapped robot's data is \textit{NOT} masked during the entire process, and the remaining robots must decide on their own (based on distances) to rely less on the data. In this experiment, we found a neighboring robot actively compensated for the missing robot and helped push the rope to the desired pose. This level of self-adaptiveness is unseen in conventional methods, and demonstrates the concept of collective intelligence in an artificial system. For further analysis, we plotted how the attention matrices changed in both trials and placed them below the corresponding screenshots in Figure~\ref{fig:self_adaptive}. Comparing (a.2) and (b.2), the attention weights to the nearest robots suddenly became larger right after kidnapping, proving that the robots are adjusting and relying more on the information from the rest of the robots to accomplish the task.

\section{CONCLUSION}

In this work, we explore incorporating the idea of collective intelligence in object manipulations by a group of cooperative mobile robots. We find that by distilling a motion planner derived from a particle-based system into an attention-based policy network, our multi-robot system is able to keep the performance of the teacher planner, and at the same time, demonstrate strong generalization capability to both unseen physics properties and the change of the number of robots. The attention matrix in the policy network allows partial interpretation of the robots behaviors, where we observe critical pattern changes at the moment when external turbulence is applied. Most importantly, the system is able to adjust and complete the task in the face of such turbulence, fully demonstrating the self-organized behaviors commonly seen in natural systems.
One of our method's limitations lies in learning efficiency, especially when the goal configurations are diverse, and we need more data for policy distillation to improve its performance.
In the future, we wish to introduce more complex tasks to further challenge our system but at the same time balance the learning efficiency. 


\printbibliography
\clearpage

\end{document}